\pdfoutput=1

\documentclass[11pt]{article}
\usepackage{emnlp2023}

\usepackage{times}
\usepackage{latexsym}

\usepackage[T1]{fontenc}

\usepackage[utf8]{inputenc}

\usepackage{microtype}
\usepackage{amsmath}

\usepackage{listings}

\usepackage{times}
\usepackage{latexsym}
\usepackage{multirow, tabu}
\usepackage{float}
\usepackage{graphicx}
\usepackage{subcaption}
\usepackage{todonotes}

\usepackage{placeins}\usepackage{xcolor}

\usepackage{arydshln}

\title{STEER: Semantic Turn Extension-Expansion Recognition for Voice Assistants}

\author{Leon Liyang Zhang\thanks{\enspace Equal contribution}\ \,, Jiarui Lu\footnote[1]{}\ \,, Joel Ruben Antony Moniz\footnote[1]{}\ \,, Aditya Kulkarni, \\
\textbf{Dhivya Piraviperumal, Tien Dung Tran, Nicholas Tzou, Hong Yu} \\
  Apple \\
\texttt{\{leonliyang\_zhang,\enspace jiarui\_lu,\enspace jramoniz,\enspace aditya,} \\
  \texttt{dhivyaprp,\enspace dung\_tran,\enspace ntzou,\enspace hong\_yu\}@apple.com} \\}

\begin{document}
\maketitle
\begin{abstract}
In the context of a voice assistant system, steering refers to the phenomenon in which a user issues a follow-up command attempting to direct or clarify a previous turn. We propose STEER, a steering detection model that predicts whether a follow-up turn is a user's attempt to steer the previous command. Constructing a training dataset for steering use cases poses challenges due to the cold-start problem. To overcome this, we developed heuristic rules to sample opt-in usage data, approximating positive and negative samples without any annotation. Our experimental results show promising performance in identifying steering intent, with over 95\% accuracy on our sampled data. Moreover, STEER, in conjunction with our sampling strategy, aligns effectively with real-world steering scenarios, as evidenced by its strong zero-shot performance on a human-graded evaluation set. In addition to relying solely on user transcripts as input, we introduce STEER+, an enhanced version of the model. STEER+ utilizes a semantic parse tree to provide more context on out-of-vocabulary words, such as named entities that often occur at the sentence boundary. This further improves model performance, reducing error rate in domains where entities frequently appear, such as messaging. Lastly, we present a data analysis that highlights the improvement in user experience when voice assistants support steering use cases.

\end{abstract}

\section{Introduction}

In the context of voice assistants, steering refers to the phenomenon in which a user issues a follow-up command attempting to direct or clarify a previous turn. However, the current state of voice assistants poorly supports steering, resulting in users having to restate their requests, causing disruptions in the natural flow of conversation and leading to a bad user experience. Support for steering use cases in voice assistants enables users to provide unprompted follow-ups, clarifying or refining their previous requests; Listing 1 presents several examples of steering use-cases.

\begin{lstlisting}[caption={Steering use case examples},label={lst:steering use case examples}]
Request                            Steering
----------------------------------------
Set an alarm at 7                  AM
Call Mom                           on Speaker
Take me to San Jose                Costa Rica
\end{lstlisting}

Building a training dataset around steering use cases is challenging because they constitute a relatively minor fraction of user follow-up requests. This is primarily due to the cold-start problem, where voice assistants poorly support steering, in turn causing users to avoid its use. Moreover, simulating training examples for steering is difficult, as arbitrarily cutting a sentence may not accurately capture the natural points in a request where users typically steer. To address this challenge, we developed heuristic rules to sample opt-in usage data without the need for explicit labeling.

As a step towards solving the under-explored problem of steering with the help of the data sampled using our heuristics, we first introduce STEER, a simple transformer-based model that utilizes query transcripts. In addition, we propose STEER+, a model that incorporates semantic parse tree (SPT) as a supplementary text-based modality. The SPT contains essential information about the intent, targets, and entities. It enhance the model's accuracy across many domains, especially in domains where entities are prevalent such as messaging. Finally, we present data analysis highlighting how support for steering use case in voice assistants can reduce user friction and improve conversation naturalness.

\section{Related Work}

\textbf{Endpoint detection} is a fundamental task in Automatic Speech Recognition (ASR) \cite{1163642, zhang2020overview}. Traditional endpoint systems use mainly acoustic information to detect where the endpoints happen \cite{li2001robust, li2002robust, yamamoto2006robust, roy2019precise}. More recently, semantic information has also been explored for the problem \cite{hwang2020end, liang2022dynamic}. All these papers work on improving the end pointer system to improve overall accuracy. On the other hand, our paper focuses on improving the end to end user experience by identifying users attempts to steer, possibly when end pointing fails.

\textbf{Sentence boundary detection}, or punctuation restoration, is a post-processing process after ASR to decide where sentences begin and end \cite{sanchez-2019-sentence, che2016sentence}. Acoustic information such as pause, pitch and speaker switch \cite{6423471, Levy2012TheEO, sinclair2014semi} and semantic information \cite{Gravano2009RestoringPA, lu2010better, ueffing2013improved, zhang2013punctuation} have been used for this problem. However, most of these methods aim to detect sentence boundaries from a long transcribed text, and assume all the previous and future text are available beforehand. Our work targets improving the understanding accuracy in a voice assistant environment where ASR transcriptions arrive in a stream, with limited future semantic context available at any given time. 

\textbf{Semantic parsing} was traditionally done using flat intent-slot schema \cite{mesnil2013investigation}. This representation was further extended to support compositional semantics using approaches like Task Oriented Parsing (TOP; \citet{gupta-etal-2018-semantic-parsing})  which represented the task in the form of a hierarchical parsing tree to allow representation for nested intents, Dialogue Meaning Representation (DMR) \cite{hu2022dialogue} that significantly extends the intent-slot framework into directed acyclic graph (DAG) composed of nodes of Intent, Entity and pre-defined Operator and Keyword, as well as edges between them.  \citet{cheng2021conversational} introduces TreeDST, which is also a tree-structured dialogue state representation to allow high compositionality and integrate domain knowledge. These complex semantic representations provide information about an ongoing task, which helps recognize if a current query is a steering of the previous one.  

\section{Motivation}

The ability to handle steering for voice assistants is crucial in enhancing the overall user experience. Firstly, it allows users to interact with voice assistants in a natural and efficient manner, without having to repeat their entire query when they want to refine or clarify a previous command. Section \ref{sec:user_friction} shows how STEER can reduce user friction from this standpoint.

Secondly, we analyze how support for steering can improve conversational naturalness in Section \ref{sec:naturalness}. In particular, support for steering allows users to pause more often, providing time for them to clarify, refine or adapt their queries through interactions. This is important in achieving more natural conversations, as humans typically have high-level intent before they speak, rather than having a fully formed query in mind. In particular, previous research has studied speech pauses in natural conversation \cite{seifart2018nouns} and in queries to voice assistants \cite{dendukuri2021using}. These studies have shown that pauses before spoken words tend to be longer when the cognitive load on the speaker is higher.

In voice assistants, balancing between latency and accuracy is an important factor in determining how long the VA system waits for a user to finish their turn \cite{Chang2022TurnTakingPF}. On one hand, a VA may shorten wait time to prioritize responsiveness; on the other, this approach could result in under-specified queries, leading to unsatisfactory responses. Steering opens up opportunities to end-point more aggressively to reduce latency while not worried about ending a request prematurely, as users can just add on to their previous command.

This requirement of responsiveness also places several limitations on the overall architecture and permissible model size and latency. Thus, although recently popular large language models (such as \citet{chung2022scaling,ouyang2022training}) are able to handle conversation end-to-end without a traditional pipelined approach (although an evaluation on how they perform on steering requests does not, to the best of our knowledge exist), they tend prohibitively expensive in terms of storage, memory, compute and inference time required, particularly if they were to be run completely on a low-power device, a setting in which voice assistants often operate.

The examples in Listing 1 highlight that queries suitable for steering are also inherently ambiguous. To leverage this observation, we propose using Semantic Parse Trees (SPTs) of a query as an additional text-based modality for modeling. These SPTs are obtained as described in \citet{cheng2021conversational,aas2023intelligent}. An example of a SPT is illustrated below in Listing~\ref{lst:sample_parse_tree}.

\begin{lstlisting}[caption={A sample Semantic Parse Tree for the request "Set an alarm at 10:30 called Bedtime"},label={lst:sample_parse_tree}]
create:alarm
    .name.Str("bedtime")
    .time.Time
        .hour.Int(10)
        .minute.Int(30)
\end{lstlisting}

When a steerable request is ambiguous, it may reflect the lack of information to construct a comprehensive SPT. This aspect can be captured by the model, facilitating more accurate steering detection. Moreover, SPTs offer a hierarchical representation of tasks, targets, and entity names parsed from a user's query, commonly used in a VA's Natural Language Understanding system \cite{cheng2021conversational}. Raw text may not reflect the presence of entity names well, since they are likely out-of-vocabulary words for a model. This issue is particularly common in steering, as named entities frequently appear at sentence boundaries, as shown in Figure~\ref{fig:part_of_speech}. Employing SPTs as a feature complements raw text by organizing the request and labeling the entities present in it.

\section{Data Sampling}
\begin{figure}[t]
    \centering
    \includegraphics[width=1\linewidth]{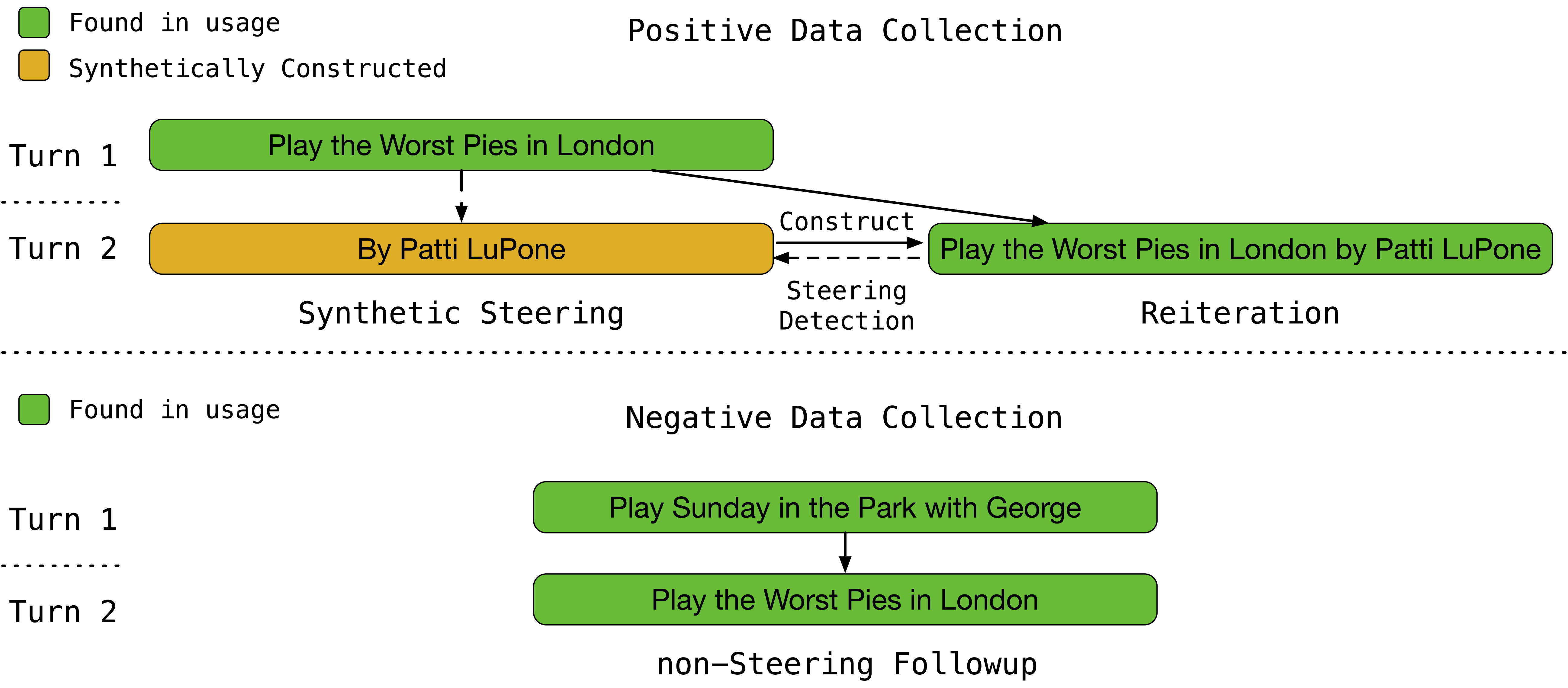}
    \caption{Illustration of data sampling process. Note that all examples shown in this paper are author-created examples based on patterns observed from anonymized and randomly sampled VA logs. In both examples, queries in green are found in real-world usage, queries in yellow are synthetically generated, representing our best guess of what the user could do if STEER is in place. For positive data, during the data sampling phase, we follow the solid lines. Given reiterations, synthetic steering follow-ups are generated. During model training, we follow the dotted lines. The model is provided with the context query and the follow-up request, then asked to predict if the follow-up is a steering request. For negative data, we use self-contained follow-ups found in the VA logs.}
    \label{fig:data}
\end{figure}

Sampling data for an unsupported task presents a challenge due to the cold-start problem: If a user attempts steering and the assistant fails to respond appropriately, the user is unlikely to attempt steering again. However, we observe that in face of an incomplete query that was incorrectly executed, users tend to reiterate the intended request in full again, resulting in a self-contained valid query. In light of this, we devised a data sampling strategy for positive data, where the follow-up intends to steer the context; and for negative data, where the follow-up is a separate request, illustrated in Fig \ref{fig:data}. Both data sampling processes start from an anonymized randomly sampled VA dataset, leveraging heuristic rules to sample opt-in usage data to create an unsupervised training set.

For positive data sampling, we first start by identifying reiterations from user. This is done by identifying consecutive turns where: 1. the previous turn is an exact prefix of the current turn; 2. the two turns happened within a short time difference. While this approach is simplistic, the resulting data is of surprisingly high quality. Next, with pairs of reiterations, we synthetically infer what a user could have said in lieu of a complete reiteration, should we have the ability to detect and handle steering. For example, in Fig \ref{fig:data}, we identified a pair of reiterations \textit{Play the Worst Pies in London} and \textit{Play the Worst Pies in London by Patti LuPone}. By extracting the suffix in the second turn, we synthetically create \textit{By Patti LuPone} as the steering follow-up for \textit{Play the Worst Pies in London}.

For negative data sampling, we capture natural non-steering follow-ups directed to the VA. As is previously mentioned, in a VA that doesn't support steering, existing steering use cases are extremely rare. Therefore, we simply sample consecutive turns from the anonymized VA usage logs of users that have opted in.

The positive and negative datasets are combined in a 1:1 ratio, resulting in a dataset of four million samples in total. This combined dataset is then randomly split into training (80\%), validation (10\%), and testing (10\%) sets. Our model is thus trained on a positive set comprising solely of unlabeled data, obtained without any annotation.

While our negative data is reflective of real-world usage, our positive dataset is derived from heuristics. To establish more certainty about this proxy dataset, we further evaluate the performance on a real-world dataset, in which steering follow-ups are manually identified and annotated from opt-in data sampled from the VA usage logs.

\section{Model}

\begin{figure*}[t]
  \centering

  \begin{subfigure}[t]{.35\linewidth}
    \centering\includegraphics[width=1\linewidth]{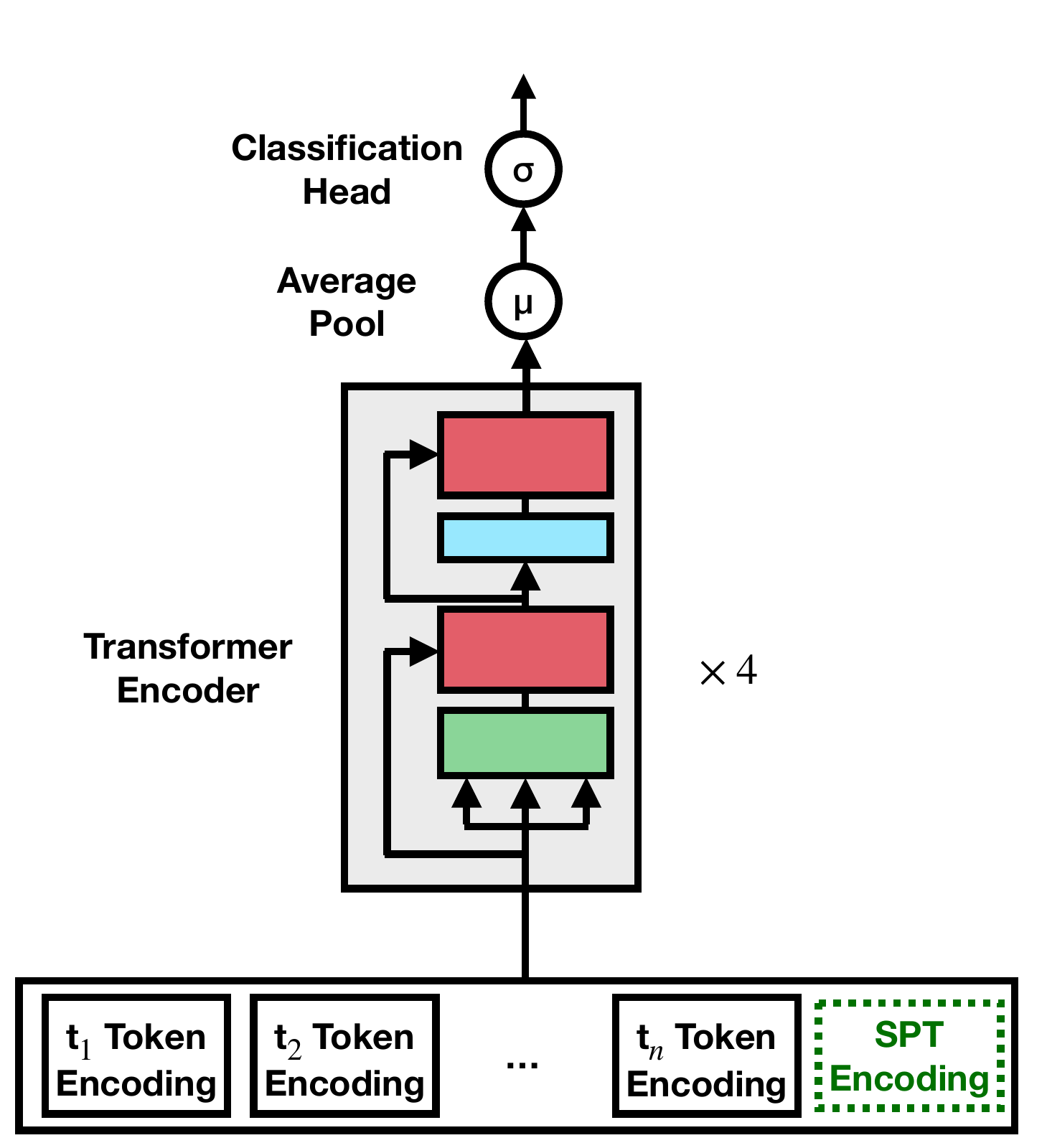}
    \caption{The overall model architecture for steering detection. Note that the concatenation of the SPT Encoding is done along the sequence dimension, rather than the hidden dimension.} \label{fig:spt_a}
  \end{subfigure}\hfill
  \begin{subfigure}[t]{.63\linewidth}
    \centering\includegraphics[width=1\linewidth]{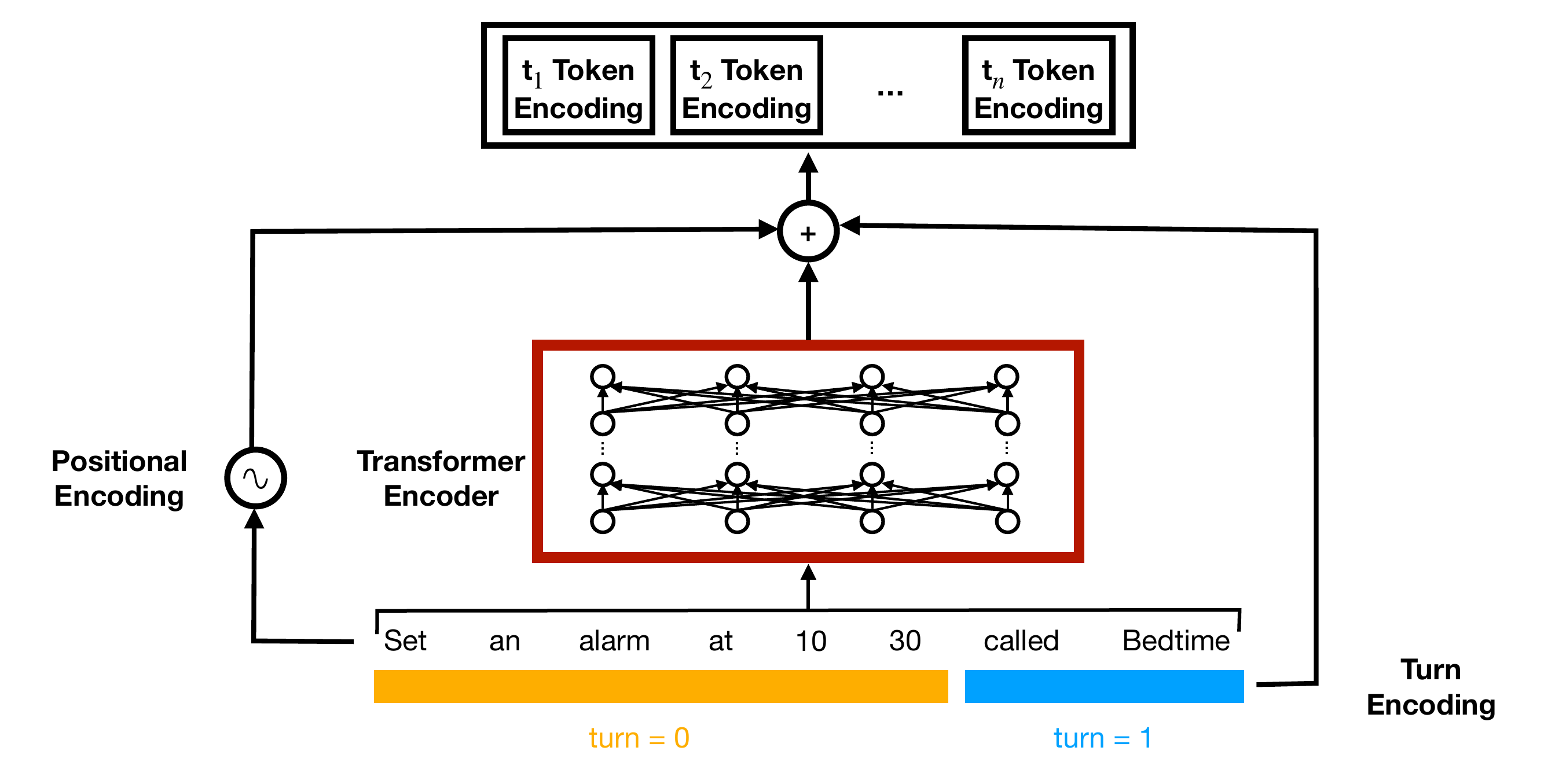}
    \caption{The encoding process for constructing Token Encodings that are inputs to the architecture in Figure~\ref{fig:spt_a}, represented by $t_1\ldots t_n$. Applies to both STEER and STEER+ architectures.} \label{fig:spt_b}
  \end{subfigure}
  \begin{subfigure}[t]{.9\linewidth}
    \centering\includegraphics[width=1\linewidth]{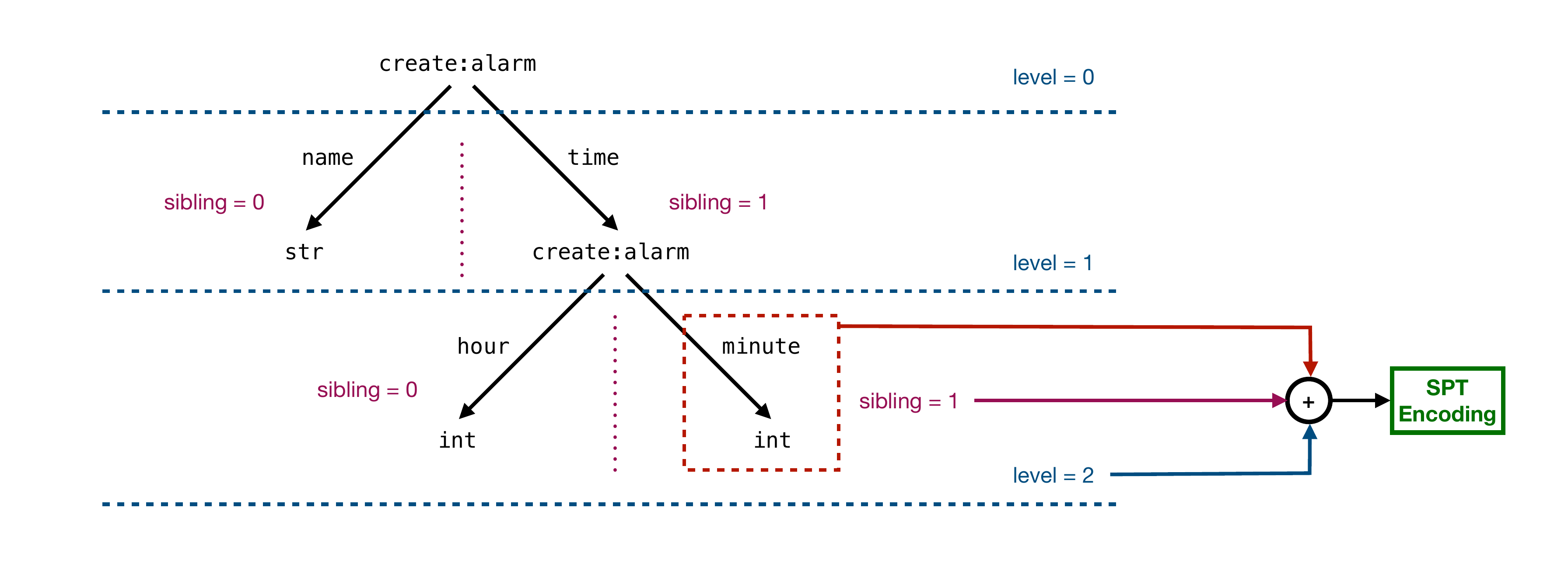}
    \caption{The encoding process of the first turn's semantic parse tree for creating SPT Encodings. This encoding is fed into Figure~\ref{fig:spt_a} for STEER+. Each SPT Encoding is the concatenation of a node, sibling, and level encoding.} \label{fig:spt_c}
  \end{subfigure}

\caption{Illustration of STEER (Figure~\ref{fig:spt_a} without SPT Encoding) and STEER+ (Figure~\ref{fig:spt_a} as a whole, including the SPT Encoding shown in dotted lines). Figures~\ref{fig:spt_b} and \ref{fig:spt_c} detail how the Token Encodings and SPT Encodings are generated respectively.}
\label{fig:spt_model}
\end{figure*}

The steering detection task can be formulated as follows: Given two turns, determine whether the second turn is a user's attempt to steer the first turn. We experimented with two variations of the transformer encoder architecture: The first model STEER, solely utilizes the transcriptions of the turns. The second model, STEER+, incorporates an additional feature: the encoding of a linearized Semantic Parse Tree (SPT) derived from the first turn.

STEER, depicted in Figure~\ref{fig:spt_model}, follows the general architecture of a transformer encoder \cite{DBLP:journals/corr/VaswaniSPUJGKP17}. It operates on tokenized queries and incorporates positional encoding and turn encoding, where the turn encoding denotes 0 for the first turn and 1 for the follow-up turn. The three encodings are projected to match the input size of the encoder and are then summed. The model consists of four transformer encoder layers, each comprising 128 hidden dimensions and 8 attention heads. Following this, the output head implements mean pooling across the sequence dimension. This pooled output is then passed through a dense classification head for the final prediction.

STEER+, also illustrated in Figure~\ref{fig:spt_model}, utilizes the same token, positional, and segment embeddings as the baseline model. However, it differs by incorporating a linearized semantic parse tree (SPT) encoding, depicted in Figure~\ref{fig:spt_c}. 
Each unique tree node, excluding payloads, is assigned a node index. To represent the SPT's structure, we introduce two additional indices: a depth index, encoding the node depth, and a sibling index, denoting the node's position among its siblings. As an example, the SPT in Listing~\ref{lst:sample_parse_tree} can be encoded as Table~\ref{tab:spt_indices}.

\begin{table}[b!]
\resizebox{\linewidth}{!}{%
\begin{tabu}{lccc}
\tabucline [1pt]{1}
Node & Node Index & Depth Index & Sibling Index \\ \hline\hline
create:alarm & v & 0 & 0 \\
.name.Str("bedtime") & w & 1 & 0 \\
.time.Time & x & 1 & 1 \\
.hour.Int(10) & y & 2 & 0 \\
.minute.Int(30) & z & 2 & 1 \\
\tabucline [1pt]{1}
\end{tabu}%
}
\caption{SPT from Listing 2 translated to indices. Nodes are encoded with indices from node vocabulary that maps to the model's encoding layer. Depth and sibling indices encode the structural information.}
\label{tab:spt_indices}
\end{table}

Once the linearized SPT is encoded into three groups of indices, we map them into three sequences of embeddings. These three SPT embeddings are then summed together, and the sum is treated as an additional token to the original query embedding along the sequence dimension. This combined input is subsequently fed into the transformer encoder and dense prediction layers, which are identical to the baseline STEER model. It is worth mentioning that we chose to keep the SPT encoding straightforward. An interesting direction of future research would be to explore more advanced techniques like Tree-LSTMs \cite{tai2015improved} and Tree Transformers \cite{nguyen2020tree} to encode the parse tree, training the system jointly in an end-to-end fashion.

\section{Experimental Setup}

Both models undergo training for a total of 300 epochs with a batch size of 256. A linear learning rate warmup is applied for the initial 30 epochs, from 1e-7 to 1e-4, followed by a linear learning rate decay throughout the remaining the epochs back to 1e-7, with early stopping. We used AdamW as model optimizer and cross-entropy as loss function.

We experimented with various training settings, such as a hyperparameter search on the learning rate and learning rate schedule, batch size variations, and other optimizers. Additionally, we explored changes in the model architecture, including varying the number of transformer encoder layers from 2 to 6, and experimenting with pooling methods such as average pooling and max pooling. Our experimental results are based on the best-performing configuration of STEER. It is important to note that our hyperparameter and architecture search was specifically done on STEER, and we maintained an identical configuration for STEER+ to ensure a fair comparison.

All our experiments were conducted on systems with a single V100 GPU. On average, STEER takes about 39 hours to train, and STEER+ trains slightly faster, for around 38 hours due to early stopping. Both STEER and STEER+ models are comparable in size, with STEER having 4.5 million parameters and STEER+ having 4.7 million.

\section{Results} 

Task performance is evaluated based on prediction accuracy on a held-out test dataset randomly split from the training data. Table~\ref{tab:results} presents the accuracies of the models, along with their 95\% confidence intervals, calculated from 32 independent trials.

STEER achieves a macro accuracy of 95.99\% $\pm$ 0.04, where macro accuracy represents the averaged classification accuracy on both data categories: consecutive reiteration (positive) and follow-ups (negative). Within each data category, STEER has an accuracy of 96.09\% $\pm$ 0.09\% on the consecutive reiteration data and 95.89\% $\pm$ 0.08\% on the follow-up data.

In comparison, STEER+ exhibits improvement over the baseline across all data categories with statistical significance. It achieves a macro accuracy of 96.44\% $\pm$ 0.03\%. Furthermore, within each data category, STEER+ attains an accuracy of 96.47\% $\pm$ 0.05\% on consecutive reiteration data and 96.40\% $\pm$ 0.06\% on follow-up data.

In addition to evaluating the models on the data collected using the heuristic sampling approach, we conducted a human grading task that involved gathering over 800 real-world steering examples. Both models were evaluated to assess their capability zero-shot in practical scenarios. Our data sampling strategy demonstrated strong alignment with real-world steering use cases, as both models achieved an accuracy of over 90\%. Moreover, STEER+ showcased statistically significantly better performance, achieving an accuracy of 91.20\% $\pm$ 0.16\%, compared to STEER with an accuracy of 90.71\% $\pm$ 0.17\%.

\begin{table}[t!]
\resizebox{\linewidth}{!}{%
\begin{tabu}{lcc}
\tabucline [1pt]{1}
   & STEER & STEER+ \\ \hline
Consecutive Reiteration Accuracy & 96.09 $\pm$ 0.09 & 96.47 $\pm$ 0.05 \\
Follow-up Accuracy & 95.89 $\pm$ 0.08 & 96.40 $\pm$ 0.06 \\ \hdashline
Macro Accuracy & 95.99 $\pm$ 0.04 & 96.44 $\pm$ 0.03 \\ \hline \hline
Real-world Positive Accuracy & 90.71  $\pm$ 0.17 & 91.20 $\pm$ 0.16 \\
\tabucline [1pt]{1}
\end{tabu}%
}
\caption{Experimental results with 95\% confidence intervals calculated from 32 independent trials. The first two rows show the accuracy for each data bucket respectively: Consecutive Reiteration data (positive) and Follow-up data (negative). The following row, macro accuracy, aggregates the two data buckets as an overall accuracy. The final row shows the accuracy of real-world graded steering use case dataset (positive).}
\label{tab:results}
\end{table}

\section{Analysis}

\begin{table}[b]
\resizebox{\linewidth}{!}{%
\begin{tabu}{lccc}
\tabucline [1pt]{1}
Domain & STEER Accuracy (\%) & STEER+ Accuracy (\%) & $\Delta$ (\%) \\ \hline\hline
Messaging & 93.54 & 96.73 & 3.18 \\
Productivity & 92.15 & 94.84 & 2.69 \\
Social Conversation & 90.92 & 92.9 & 1.98 \\
Images & 96.46 & 98.23 & 1.77 \\
Ambiguous & 93.77 & 94.75 & 0.98 \\
Web Search & 94.93 & 95.84 & 0.91 \\
Music & 97.32 & 98.11 & 0.8 \\
Sports & 96.45 & 97.16 & 0.71 \\
Phone Call & 94.08 & 94.62 & 0.54 \\
Knowledge & 95.83 & 96.26 & 0.43 \\
Video & 91.76 & 92.05 & 0.28 \\
Math & 97.91 & 98.17 & 0.26 \\ \hline
Weather & 98.39 & 98.27 & -0.12 \\
Maps & 95.75 & 95.47 & -0.28 \\
System Actions & 96.18 & 95.83 & -0.35 \\
Time Utilities & 98.14 & 97.15 & -0.99 \\
\tabucline [1pt]{1}
\end{tabu}%
}
\caption{Domain-wise break down of STEER and STEER+ performance in accuracy on 20k positive test samples. $\Delta$ highlights the performance difference of STEER+ over STEER.}
\label{tab:result_details}
\end{table}

From Table~\ref{tab:results}, we observe that incorporating SPT into our model leads to improved accuracy. Our hypothesis is that when the first turn is steerable, its corresponding SPT can be enriched with additional information, which signals incompleteness. Furthermore, steering often involves clarification with entity names as evident Figure 3, and the SPT can offer context on where these entities occur, enhancing the model's understanding. To further validate our assumption that the SPT can provide model with entity context, Table~\ref{tab:result_details} shows a domain-wise break-up of STEER and STEER+ performance. STEER+ shows significant gains in entity prevalent domains such as messaging, social conversation, and images. Fusing SPT also improves STEER+'s performance across most other domains, with only minor drops in a few specific domains.

In addition to model evaluation, we also assessed the benefits of a voice assistant system having steering support brings to end users. This analysis focus on two aspects: we first show that there is a substantial reduction in user friction. Then, we demonstrate how support for steering improves conversational naturalness by allowing users to pause and formulate (or refine) their query. We present additional analysis in Appendix~\ref{app:analysis}.

\subsection{Reducing User Friction}
\label{sec:user_friction}

To quantify how a steering-enhanced system can help reduce user friction, we design a proxy metric, which involves measuring the number of words a user is saved by not having to reiterate their entire query, since users can simply pick up where they left off in the previous turn by issuing a steering followup.

Given a steering use case, we quantify the overall user friction reduction as $f$, as outlined in equation~\ref{eqn:impact}. When the model correctly predicts steering, the user does not have to repeat the original request and can continue with the steering command, resulting in a friction reduction of $f_\text{request}$. However when the model fails to predict steering correctly, the user has already issued the steering request, leading to additional friction, denoted by $f_\text{steer}$. This indicates that the user has paid an extra cost compared to a voice assistant system that does not support steering: 

\begin{equation} \label{eqn:impact}
f = f_\text{request} \cdot \hat{y}  - f_\text{steer} \cdot (1 - \hat{y}),
\end{equation}
where $\hat{y}$ is the model's prediction.
We measure user friction $f$ in equation~\ref{eqn:impact} as the average number of words saved and average proportion of total query saved by steering as detailed in Table~\ref{tab:user_friction}.

\begin{table}[!h]
\resizebox{\linewidth}{!}{%
\begin{tabu}{lccc}
\tabucline [1pt]{1}
 & Words Saved & Fraction of Query Saved (\%) \\ \hline\hline
STEER & $3.963 \pm $0.007 & $58.06 \pm $0.07  \\
STEER+ & $4.095 \pm $0.005 & $58.64 \pm $0.05  \\
Upper Bound & 4.417 & 62.17  \\
\tabucline [1pt]{1}
\end{tabu}%
}
\caption{Reduction in user friction is compared between STEER and STEER+ on a 20k positive test set. On average, STEER saves 58.06\% of query from repetition (equivalently 3.96 words per query). 0.6\% abs improvement observed with STEER+. A perfect model (upper bound) will save 62.17\% of request from repetition.}
\label{tab:user_friction}
\vspace{-0.2\baselineskip}
\end{table}

\begin{figure}[!t]
    \centering
    \includegraphics[width=1\linewidth]{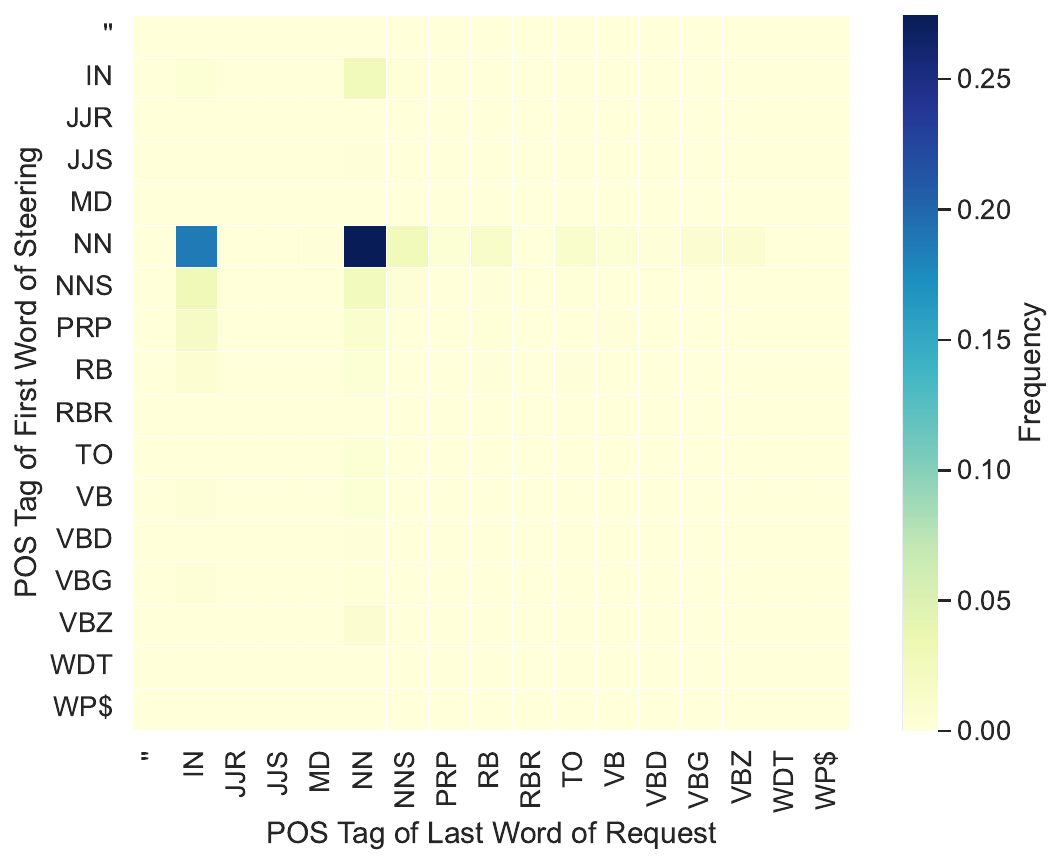}
    \caption{Illustration of part-of-speech transition probability at the steering boundary. Steering is most frequent when the user provides a named entity; in particular, transitions from prepositions/subordinating conjunctions (IN) to nouns (NN) (for example: 'what time is it in', 'portland') and NN to NN (for example: 'how far is las vegas from watsonville', 'california') are common.}
    \label{fig:part_of_speech}
    \vspace{-0.3\baselineskip}
\end{figure}

\subsection{Improving Conversation Naturalness} \label{sec:naturalness}

Steering provides the ability to handle disfluencies in user speech, which might include thought pauses and slow speech. This is expected to be more pronounced before named entities \cite{seifart2018nouns,dendukuri2021using}. Figure~\ref{fig:part_of_speech} shows steering to be robust to such disfluency in speech. Steering allows the user the flexibility to provide named entities in a separate request, and avoids the need for users to be prepared with an entire query before engaging with voice assistants. This flexibility enables fluid conversations, allowing users to have a natural, human-like experience. Since steering can be triggered multiple times within a single request, it offers support for long and complex requests. The steering explored in this work thus serves as a foundational framework for building next generation voice assistants that are capable of executing complex instructions, often involving multiple tasks. 

\section{Conclusion}
In this work, we proposed STEER, a steering detection model for voice assistants. Our research presents a data sampling strategy that enables us to obtain high quality steering data without annotation. Additionally, We introduced STEER+, which jointly learns from token features and a semantic parse tree, achieving over $91\%$ classification accuracy on real-world data, and showing significant error reduction and lower user friction over STEER. Lastly, we present a data analysis highlighting how support for steering use case in voice assistants can reduce user friction and improve conversation naturalness. We hope that this work can support future research and advancements in VA systems, ultimately enhancing their capabilities and usability in various domains.

\FloatBarrier

\section*{Acknowledgements}

The authors would like to thank Murat Akbacak, Daryn Bryden, Anu Gali, Zong-Fu Hsieh, Woojay Jeon, Xiaochuan Niu, Nidhi Rajshree, Ahmed Tewfik, Bo-Hsiang Tseng and the anonymous reviewers for their help and feedback.

\bibliography{anthology,custom}
\bibliographystyle{acl_natbib}

\clearpage

\appendix

\section{Additional Analysis} \label{app:analysis}

In this appendix section, we present additional analysis around the impact of using SPTs in STEER+ and the benefits this offers over STEER. We also delve deeper into the reduction in user friction explored in Section~\ref{sec:user_friction}, both by examining our proposed STEER model, but also by examining STEER+ in comparison with STEER.

\subsection{STEER vs STEER+}

We further dive the comparison between STEER and STEER+, visualizing the statistically significant improvement provided from encoding the SPT in STEER+ demonstrated in Table~\ref{tab:results}.

\begin{figure}[h]
  \centering

  \begin{subfigure}[t]{.95\linewidth}
    \centering\includegraphics[width=1\linewidth]{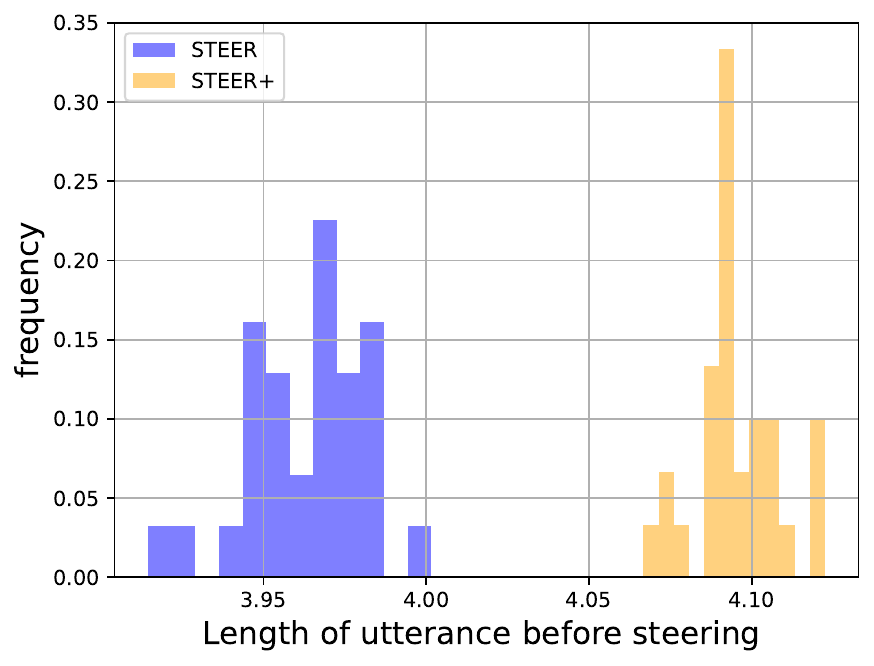}
    \caption{} \label{fig:runs_a}
  \end{subfigure}\hfill
  \begin{subfigure}[t]{.95\linewidth}
    \centering\includegraphics[width=1\linewidth]{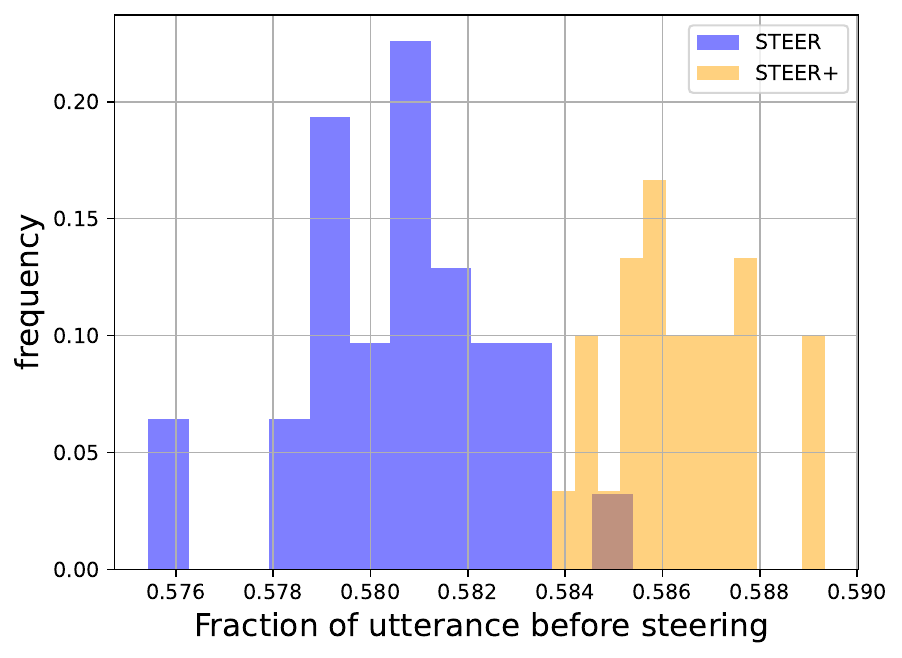}
    \caption{} \label{fig:runs_b}
  \end{subfigure}

\caption{STEER+ shows statistically significant improvement over STEER as observed by the distributions above and 95\% CI reported in Table~\ref{tab:user_friction}. This highlights the benefit of using semantic signals to the end user.}
\label{fig:runs}
\end{figure}

In Figure~\ref{fig:runs} above shows histograms that aim to summarize the performance of all STEER and STEER+ models trained in terms of how they help reduce user friction (refer Section~\ref{sec:user_friction} for details). In sub-plot~\ref{fig:runs_a}, the x-axis buckets represent the (absolute) number of words saved when a steering system is in place; in sub-plot~\ref{fig:runs_b} the x-axis buckets represent the fraction of the query that the user does not have to repeat. In both sub-plots, the y-axis captures how many among our repeated, independent trials fell into a particular bucket. 

As is evident from the figure, in both cases, we see that even the worst performing STEER model helps reduce user friction. Interestingly, as is seen from the two histograms, we find that even the best performing STEER model from all our runs is comparable to among the worst performing STEER+ models, with a clear separation between the two histograms in sub-plot~\ref{fig:runs_a} and almost no overlap in sub-plot~\ref{fig:runs_b}.

\subsection{User Impact Breakdown}

To further explore the user impact and reduction in user friction analyzed in Section~\ref{sec:user_friction}, we present two histograms below. 

\begin{figure}[h]
    \centering
    \includegraphics[width=1\linewidth]{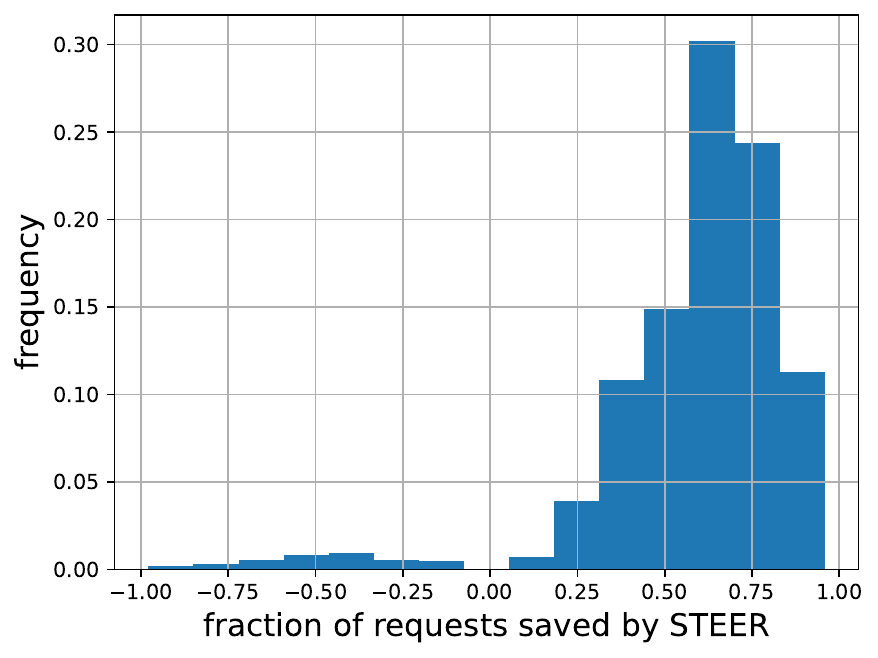}
    \caption{Illustration of user impact from steering. In some instances, when the model fails to detect steering, the user has to repeat their request on top of the query that they used to attempt to steer: this accounts for the distribution on the negative side of the x-axis. However, since most steering cases are detected correctly, the STEER model effectively provides a significant net benefit to the user.}
    \label{fig:fraction_of_request_saved_by_steer}
\end{figure}

Figure~\ref{fig:fraction_of_request_saved_by_steer} is a histogram in which the buckets on the x-axis represent fraction of the requests saved by STEER, as calculated by Equation~\ref{eqn:impact}; the y-axis represents the frequency (as a fraction of our analysis set). While there are a very small number of cases where STEER degrades the user's experience by incorrectly identifying a steering request as a follow-up, the figure clearly illustrates that, an overwhelming majority of the time, there is a net improvement in the user experience.

\begin{figure}[h]
    \centering
    \includegraphics[width=1\linewidth]{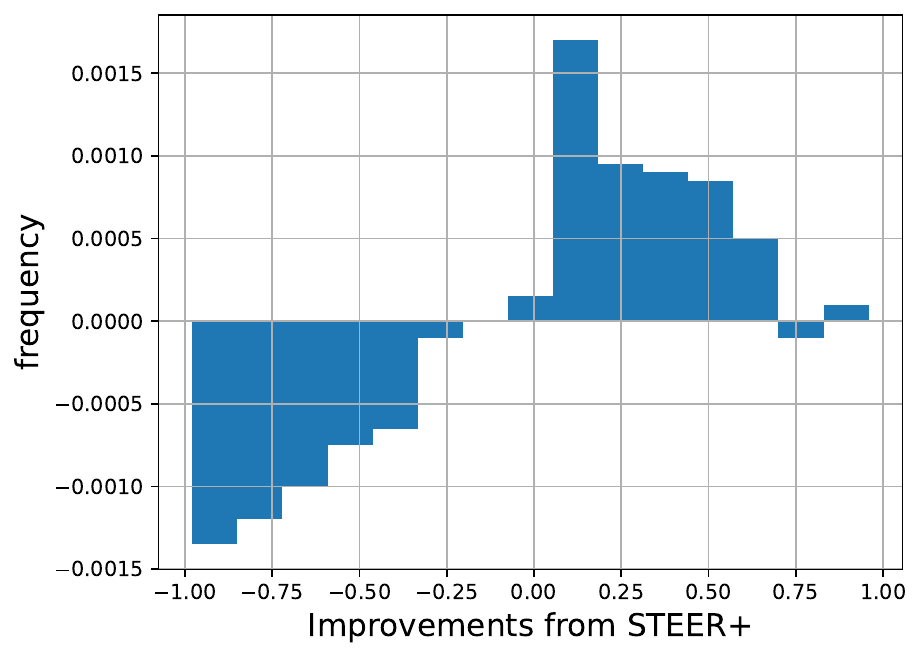
    }
    \caption{Illustration of improvement from STEER+. Owing to the improved accuracy of STEER+, we see better detection of steering. This, in turn, results in fewer instances of users repeating the steering query. We thus see an improvement in terms of user experience by incorporating semantic signals}
    \label{fig:Improvements_from_STEER+}
\end{figure}

Figure~\ref{fig:Improvements_from_STEER+} aims to show a detailed comparison of how STEER+ performs in comparison to STEER. To do this, as in Figure~\ref{fig:fraction_of_request_saved_by_steer}, we show a histogram in which the y-axis represents the frequency (as a fraction of our analysis set); however, here, the buckets on the x-axis represent the \emph{difference} in the fraction of the requests saved by STEER and those saved by STEER+. A positive value means that STEER+ has a higher frequency of datapoints falling into that bin than STEER. Here, we see that almost all bins below 0 are negative, which implies that STEER+ consistently \emph{reduces} the number of cases in which steering failed to be detected; likewise, almost all bins above 0 are positive, implying that STEER+ consistently \emph{increases} the number of cases in which steering was correctly detected.

\end{document}